\documentclass[a4paper]{llncs}

\usepackage[utf8]{inputenc}
\usepackage[english]{babel}
\usepackage{amsmath}
\usepackage{amsfonts}
\usepackage{float}

\floatstyle{plaintop}
\restylefloat{table}

\usepackage{subfig}
\usepackage{graphicx}
\usepackage{authblk}
\usepackage{color}
\usepackage{rotating}
\usepackage{makecell}
\usepackage{cite}
\usepackage[misc]{ifsym}

\usepackage[hyphens]{url}
\usepackage[hidelinks]{hyperref}
\hypersetup{breaklinks=true}
\graphicspath{{figure/}}

\title{FastVentricle: Cardiac Segmentation with ENet}
\author{Jesse Lieman-Sifry$^{\textrm{\Letter}}$, Matthieu Le, Felix Lau, Sean Sall, and Daniel Golden}
\institute{Arterys, San Francisco, USA \\  \email{first@arterys.com}}

\begin{document}
\maketitle

\begin{abstract} 
Cardiac Magnetic Resonance (CMR) imaging is commonly used to assess cardiac structure and function. One disadvantage of CMR is that post-processing of exams is tedious. Without automation, precise assessment of cardiac function via CMR typically requires an annotator to spend tens of minutes per case manually contouring ventricular structures. Automatic contouring can lower the required time per patient by generating contour suggestions that can be lightly modified by the annotator. Fully convolutional networks (FCNs), a variant of convolutional neural networks, have been used to rapidly advance the state-of-the-art in automated segmentation, which makes FCNs a natural choice for ventricular segmentation. However, FCNs are limited by their computational cost, which increases the monetary cost and degrades the user experience of production systems. To combat this shortcoming, we have developed the FastVentricle architecture, an FCN architecture for ventricular segmentation based on the recently developed ENet architecture. FastVentricle is $4\times$ faster and runs with $6\times$ less memory than the previous state-of-the-art ventricular segmentation architecture while still maintaining excellent clinical accuracy.
\end{abstract}

\section{Introduction}

Patients with known or suspected cardiovascular disease often receive a cardiac MRI to evaluate cardiac function. These scans are annotated with ventricular contours in order to calculate cardiac volumes at end systole (ES) and end diastole (ED); from the cardiac volumes, relevant diagnostic quantities such as ejection fraction and myocardial mass can be calculated.  Manual contouring can take upwards of 30 minutes per case, so radiologists often use automation tools to help speed up the process. 

\begin{figure}
  \centering
  \includegraphics[width=0.7\linewidth]{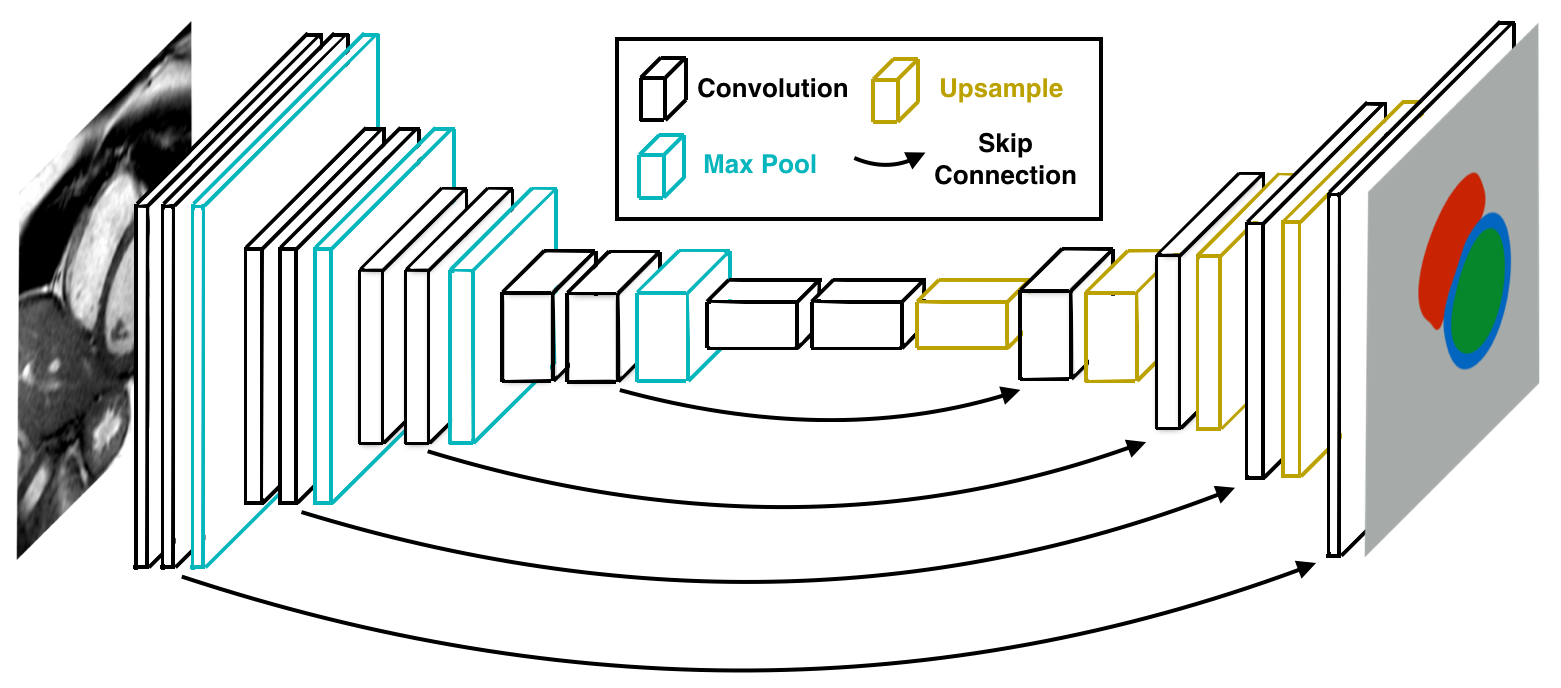}
  \caption{Schematic representation of a fully convolutional encoder-decoder architecture with skip connections that utilizes a smaller expanding path than contracting path.}
  \label{fig:fcn}
\end{figure}

Active contour models \cite{kass1988} are a heuristic-based approach to segmentation that have been utilized previously for segmentation of the ventricles \cite{choa2006,zhu2013geodesic} with optional use of a ventricle shape prior \cite{plue2005, schwarz20073d}. However, active contour-based methods not only perform poorly on images with low contrast, they are also sensitive to initialization and hyperparameter values. We encourage the interested reader to refer to recent review papers \cite{petitjean2011review,peng2016review} as a jumping-off point for further insight on the usage of these (and many other) non-deep learning approaches for cardiac segmentation.

Deep learning methods for segmentation have recently defined state-of-the-art with the use of fully convolutional networks (FCNs) \cite{long2015fully}. Simple FCN architectures similar to that described by \cite{long2015fully} have been utilized for cardiac segmentation \cite{tran2016fully}. The general idea behind FCNs is to use a downsampling path to learn relevant features at a variety of spatial scales followed by an upsampling path to combine the features for pixelwise prediction (see Fig. \ref{fig:fcn}). DeconvNet \cite{noh2015learning} pioneered the use of a symmetric contracting-expanding architecture for more detailed segmentations, at the cost of longer training and inference time, and the need for larger computational resources. UNet \cite{ronneberger2015u}, originally developed for use in the biomedical community where there are often fewer training images and even finer resolution is required, added the use of skip connections between the contracting and expanding paths to preserve details. 

A UNet variant, DeepVentricle, has previously been used for cardiac segmentation \cite{lau2016} and has received FDA clearance for clinical usage \cite{arterysfda}. Both UNet and DeepVentricle are FCNs with symmetric downsampling and upsampling paths, skip connections, two convolution operations before each pooling or upsampling operation, and double/half the number of filters as in the previous layer following each pool/upsample, respectively. A key difference is that UNet utilizes valid padding, resulting in segmentation maps that are smaller than the input image, whereas DeepVentricle uses same padding.

One disadvantage of fully symmetric architectures in which there is a one-to-one correspondence between downsampling and upsampling layers is that they can be slow, especially for large input images. ENet, an alternative FCN design, is an asymmetrical architecture optimized for speed \cite{paszke2016enet}. ENet utilizes early downsampling to reduce the input size using only a few feature maps. This reduces both training and inference time, given that much of the network's computational load takes place when the image is at full resolution, and has minimal effect on accuracy since much of the visual information at this stage is redundant. Furthermore, the ENet authors show that the primary purpose of the expanding path in FCNs is to upsample and fine-tune the details learned by the contracting path rather than to learn complicated upsampling features; hence, ENet utilizes an expanding path that is smaller than its contracting path. ENet also makes use of bottleneck modules, which are convolutions with a small receptive field that are applied in order to project the feature maps into a lower dimensional space in which larger kernels can be applied \cite{he2016deep}. Finally, throughout the network, ENet leverages a diversity of low cost convolution operations. In addition to the more-expensive $n \times n$ convolutions, ENet also uses cheaper asymmetric ($1 \times n$ and $n \times 1$) convolutions and dilated convolutions \cite{yu2015multi}. 

In this paper, we present FastVentricle, an ENet variation with skip connections for segmentation of the LV endocardium (LV endo), LV epicardium (LV epi), and RV endocardium (RV endo), and compare it to DeepVentricle, the architecture previously cleared by the FDA for clinical use. We show that inference with FastVentricle requires significantly less time and memory than inference with DeepVentricle while achieving statistically indistinguishable volume accuracy. 

\section{Methods}

\noindent
\textbf{Training Data.} We use a database of 1143 short-axis cine Steady State Free Precession (SSFP) scans annotated as part of standard clinical care to train and validate our model. We split the data chronologically with 80\% of studies used for training, 10\% for validation, and 10\% as a hold out set. We curate the hold out set to only include contours from trusted annotators, i.e. licensed radiologists or technologists. The hold out set is used for all plots and tables in this paper. Annotated contour types include LV endo, LV epi and RV endo. Scans are annotated at ED and ES. Contours were annotated with different frequencies; 96$\%$ (1097) of scans have LV endo contours, 22$\%$ (247) have LV epi contours and 85$\%$ (966) have RV endo contours.

\textbf{Data Preprocessing.} We normalize all MRIs such that the 1st and 99th percentile of pixel intensities of a batch of images fall at -0.5 and 0.5, i.e. their “usable range” falls between -0.5 and 0.5. We crop and resize the images such that the ventricle contours take up a larger percentage of the image; the actual crop and resize factors are hyperparameters. Cropping the image increases the fraction of the image that is taken up by the foreground (ventricle) class, making it easier to resolve fine details and helping the model converge. 

\textbf{Training.} We use the Keras \cite{chollet2015keras} deep learning package with TensorFlow \cite{abadi2016tensorflow} as the backend to implement and train all of our models. We modify the standard per-pixel cross-entropy loss to account for missing ground truth annotations in our dataset. We discard the component of the loss that is calculated on images for which ground truth is missing; we only backpropagate the component of the loss for which ground truth is known. This allows us to train on our full training dataset, including series with missing contours. Network weights are updated per the Adam rule \cite{kingma2014adam}. We train and test the models using NVIDIA Maxwell Titan X GPUs. We augment our data during training by flipping, shearing, shifting, zooming, and rotating the image/label pairs. To compare different models, we use relative absolute volume error (RAVE). Using a relative metric ensures that equal weight is given to pediatric and adult hearts. We focus on the volume error, as opposed to the Sørensen–Dice index or a similar overlap metric, because ventricular volumes are the clinical endpoint used to diagnose disease and determine clinical care. RAVE is defined as $\text{RAVE} = |\text{V}_{pred} - \text{V}_{truth}| / \text{V}_{truth}$, where $\text{V}_{truth}$ is the ground truth volume and $\text{V}_{pred}$ is the volume computed from the predicted 2D segmentation masks. Volumes are calculated from segmentation masks using a frustum approximation.

\begin{figure}[t]
	\centering
    \hspace{\fill} 
    \includegraphics[width = \linewidth]{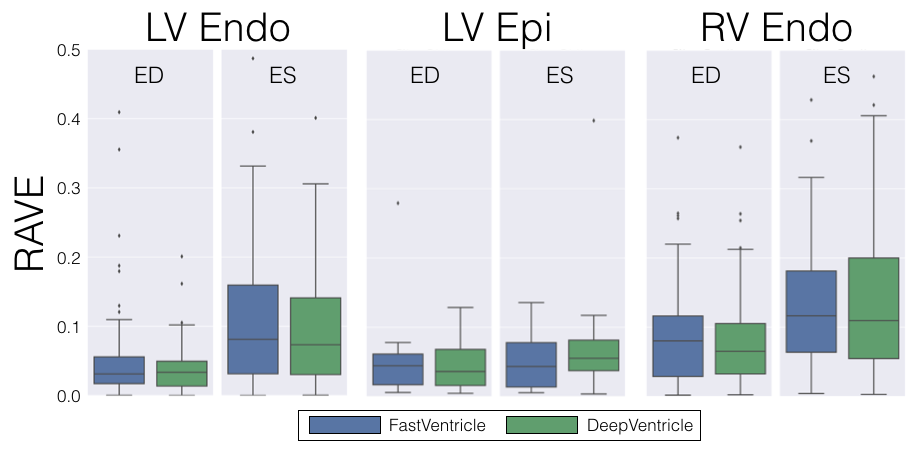}
	\hspace{\fill} 
    \caption{Boxplots comparing the relative absolute volume error (RAVE) between FastVentricle and DeepVentricle for each of LV endo, LV epi, and RV endo at ED (left panels) and ES (right panels). The line at the center of each box denotes the median RAVE, the ends of the box show 25$\%$ (Q1) and 75$\%$ (Q3) of the distribution. Whiskers are placed at the first value within 1.5 interquartile ranges past the first and third quartiles.}
    \label{fig:inference_comparison}
\end{figure}

\textbf{Hyperparameter Search.} We fine tune the DeepVentricle and FastVentricle network architectures using random hyperparameter search \cite{bergstra2012random}. In practice, for each of the DeepVentricle and FastVentricle architectures, we: i) run models with random sets of hyperparameters for 20 epochs (i.e. 20 passes over the training set), ii) select from the resulting corpus of models the three models with the highest validation set accuracy, iii) select the final model from the three candidates based on lowest average RAVE on the validation set. The hyperparameters of the DeepVentricle architecture include the use of batch normalization, the number of convolution layers, the number of initial filters, and the number of pooling layers. The hyperparameters of the FastVentricle architecture include the kernel size for asymmetric convolutions, the number of times Section 2 of the network is repeated, the number of initial bottleneck modules, the number of initial filters, the projection ratio, and whether to use skip connections. These connections run from the contracting path (the initial block and Section 1) to the equivalent image size on the expanding path (Section 5 and Section 4, respectively). Refer to \cite{paszke2016enet} for nomenclature of the Sections and detailed descriptions of the hyperparameters. For both architectures, the hyperparameters also include the batch size, learning rate, dropout probability, crop fraction, image size, and the strength of all data augmentation parameters. We trained 50 DeepVentricle models and 20 FastVentricle models for our hyperparameter search, with the scope of the search limited by computational limitations and time constraints. We note that skip connections are used in the 5 best FastVentricle architectures in terms of validation set accuracy, demonstrating their usefulness for this problem.

\begin{table}[b]
  \centering
  \resizebox{\textwidth}{!}{\begin{tabular}{|l|c|c|c|c|c|c|}
    \hline
      & LV Endo (ED) & LV Endo (ES) & LV Epi (ED) & LV Epi (ES) & RV Endo (ED) & RV Endo (ES) \\
    \hline
    \makecell{DeepVentricle \\ Median RAVE} & 0.033 & 0.073 & 0.036 & 0.049 & 0.064 & 0.109 \\
    \hline
    \makecell{FastVentricle \\ Median RAVE} & 0.031 & 0.081 & 0.044 & 0.043 & 0.080 & 0.116 \\      
    \hline
     U statistic & 4678 & 4967 & 190 & 160 & 4623 & 4142 \\
    \hline
     p-value & 0.86 & 0.35 & 0.79 & 0.56 & 0.28 & 0.80 \\
    \hline
     Sample size & 96 & 96 & 19 & 19 & 112 & 112 \\ 
    \hline
  \end{tabular}}
  \caption{Median RAVEs, U statistics and p-values from the Wilcoxon-Mann-Whitney test, and corresponding sample sizes for our comparison of DeepVentricle and FastVentricle for every combination of phase and ventricular anatomy on our hold out set. We observe no statistically significant difference between the RAVE distributions of DeepVentricle and FastVentricle.}
  \label{table:statistical_sig}
\end{table}

\section{Results}

\begin{sidewaysfigure}
    \centering
	\includegraphics[width = \linewidth]{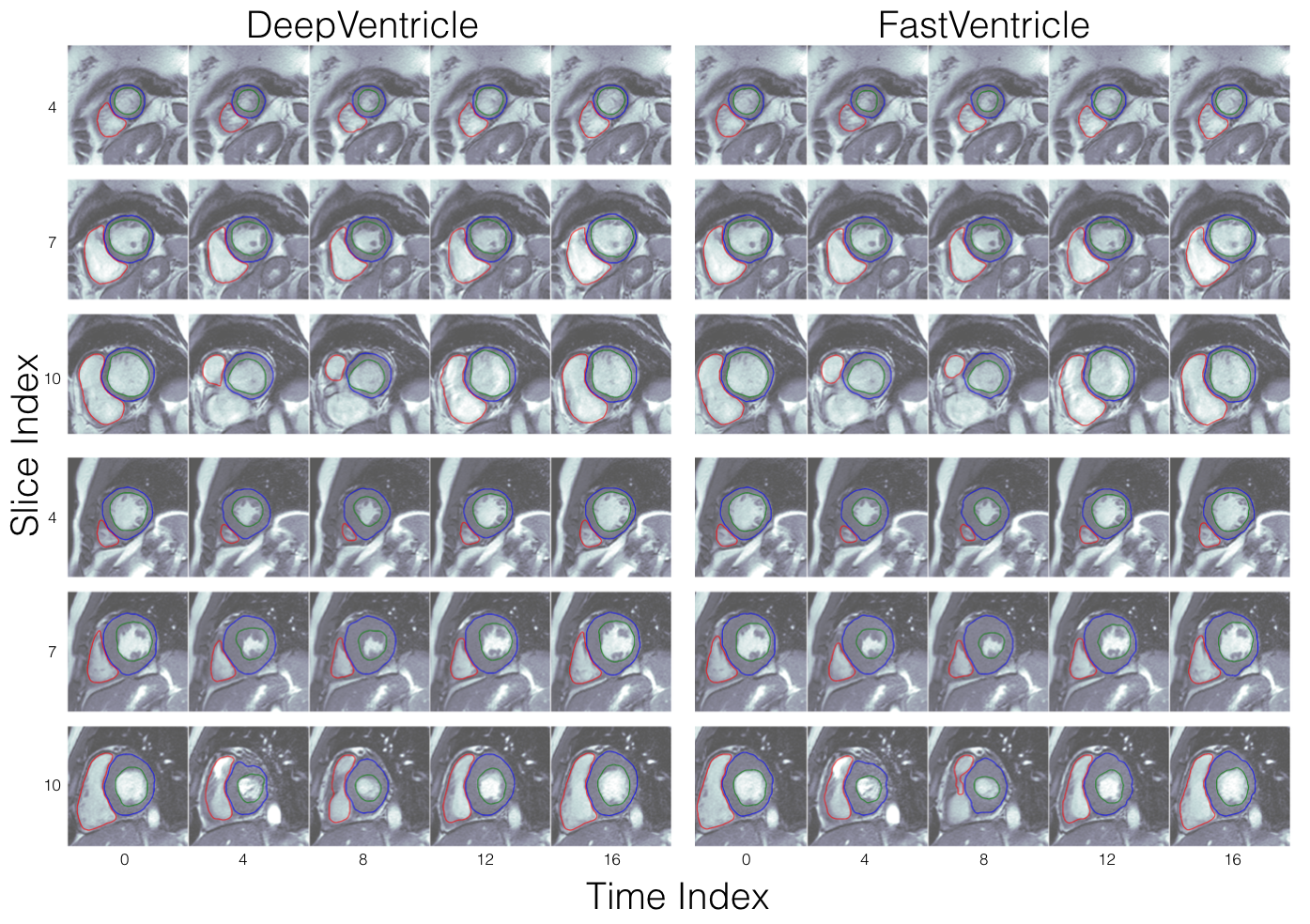} 
	\caption{DeepVentricle and FastVentricle predictions for a healthy patient (top) and on a patient with hypertrophic cardiomyopathy (HCM, bottom). RV endo is outlined in red, LV endo in green, and LV epi in blue. DeepVentricle's average RAVE is 0.080 and 0.095 for the healthy and HCM patients, respectively, and FastVentricle's average RAVE is 0.057 and 0.053 for each patient, respectively. The x-axis of the grid corresponds to time indices sampled throughout the cardiac cycle and the y-axis corresponds to slice indices sampled from apex (low slice index) to base (high slice index).}
    \label{fig:enet_good_inference}
\end{sidewaysfigure}

\textbf{Volume Error Analysis.} 
Figure \ref{fig:inference_comparison} presents box plots of the RAVE comparing DeepVentricle and FastVentricle for each combination of ventricular structure (LV endo, LV epi, RV endo) and phase (ES, ED). Within our hold out set, we find that the performance of the models are very similar across structures and phases (see Table \ref{table:statistical_sig} for median RAVEs as well as corresponding sample sizes). For both models, segmentations at ED tend to be better than at ES, as the chambers at ES are smaller and dark-colored papillary muscles tend to blend with the myocardium when the heart is contracted. RV endo is the most difficult of the structures due to its more complex shape. Furthermore, we find that model performance at the apex and center of the ventricle is often better than at the base, as it is generally ambiguous from just the basal slice where the valve plane (separating ventricle from atrium) is. Although trained on only ES and ED annotations, we are able to perform visually pleasing inference on all time points. Figure \ref{fig:enet_good_inference} shows examples of network predictions on different slices and time points for studies for both DeepVentricle and FastVentricle. 

\begin{figure}
    \centering
	\includegraphics[width = \linewidth]{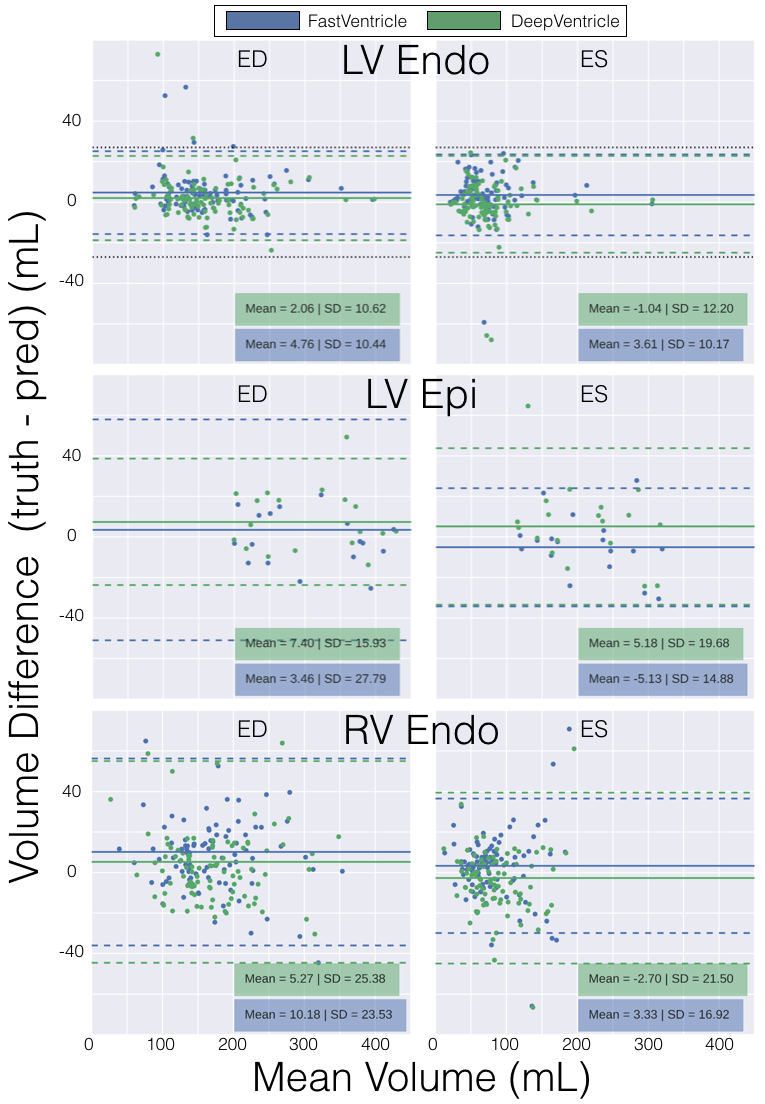} 
	\caption{Bland-Altman plots of ventricle volumes for both DeepVentricle (green) and FastVentricle (blue) for each combination of ventricular structure and phase vs. ground truth annotations. Solid colored lines denote the mean difference between ground truth and predicted volumes, and dashed colored lines show the mean $\pm 1.96SD$, where $SD$ is the standard deviation of the difference. The mean can be interpreted as the bias of the new method (automatic segmentation) vs. the old (doctor's annotations), and the dashed lines are the 95$\%$ limits of agreement between the two methods. For reference, \cite{suinesiaputra2015quantification} find that the 95$\%$ limits of agreement between expert physicians is approximately $\pm 27$mL for LV endo measurements; we display this value as a dotted black line on the LV endo plots. Our LV endo volume 95$\%$ limits of agreement is within that of the 95$\%$ limits of agreement between expert annotators for both DeepVentricle and FastVentricle. Information on the 95$\%$ limits of agreement between expert annotators is unavailable for LV epi and RV endo.}
    \label{fig:ba_volumes}
\end{figure}

We additionally make Bland-Altman plots \cite{bland1986statistical} for the volume error for each combination of ventricular structure and phase (Figure \ref{fig:ba_volumes}). These plots are used to show the agreement between the ``gold standard" method (i.e. expert manual annotations) and our automated DeepVentricle and FastVentricle methods. If the 95$\%$ limits of agreement (mean $\pm 1.96 SD$) are within the range that would not make a clinical difference, the new method can be used in place of the old with no adverse effects. 

\textbf{Statistical Analysis.} We measure the statistical significance of the difference between DeepVentricle and FastVentricle's RAVE distributions for each combination of phase and anatomy for which we have ground truth annotations. We use the Wilcoxon-Mann-Whitney test \footnote{Using the SciPy 0.17.0 implementation with default parameters https://docs.scipy.org/doc/scipy/reference/generated/scipy.stats.mannwhitneyu.html} to assess the null hypothesis $H_0$ that the distribution of DeepVentricle and FastVentricle's RAVE are equal. Table \ref{table:statistical_sig} displays the results. We find that there is no statistical evidence to claim one model as the best, since the lowest measured p-value is 0.28.

\textbf{Computational Complexity and Inference Speed.} To be clinically and commercially viable, any automated algorithm must be faster than manual annotations, and lightweight enough to deploy easily. As seen in Table \ref{table:model_comparison}, we find that FastVentricle is roughly $4\times$ faster than DeepVentricle and uses $6\times$ less memory for inference. Because the model contains more layers, FastVentricle takes longer to initialize before being ready to perform inference. However, in a production setting, the model only needs to be initialized once when provisioning the server, so this additional cost is incidental.

\begin{table}[b]
  \centering
  \begin{tabular}{|l|c|c|}
    \hline
      & DeepVentricle & FastVentricle \\
    \hline
     Average RAVE & 0.100 & \textbf{0.098} \\
    \hline
     Median RAVE & \textbf{0.057} & 0.065 \\     
    \hline
     Inference GPU time per sample (msec) & 31 & \textbf{7} \\
     \hline
     Initialization GPU time (sec) & \textbf{1.3} & 13.3 \\
    \hline
     Number of parameters & 19,249,059 & \textbf{755,529} \\
    \hline
     GPU memory required for inference (MB) & 1,800 & \textbf{270} \\
    \hline
     Size of the weight file (MB) & 220 & \textbf{10} \\
    \hline
  \end{tabular}
  \caption{Model volume error, speed, and computational complexity for DeepVentricle and FastVentricle. Inference time per sample and GPU memory required for inference calculated with a batch size of 16.}
  \label{table:model_comparison}
\end{table}

\begin{figure}[t]
    \centering
	\subfloat{\includegraphics[width = \linewidth]{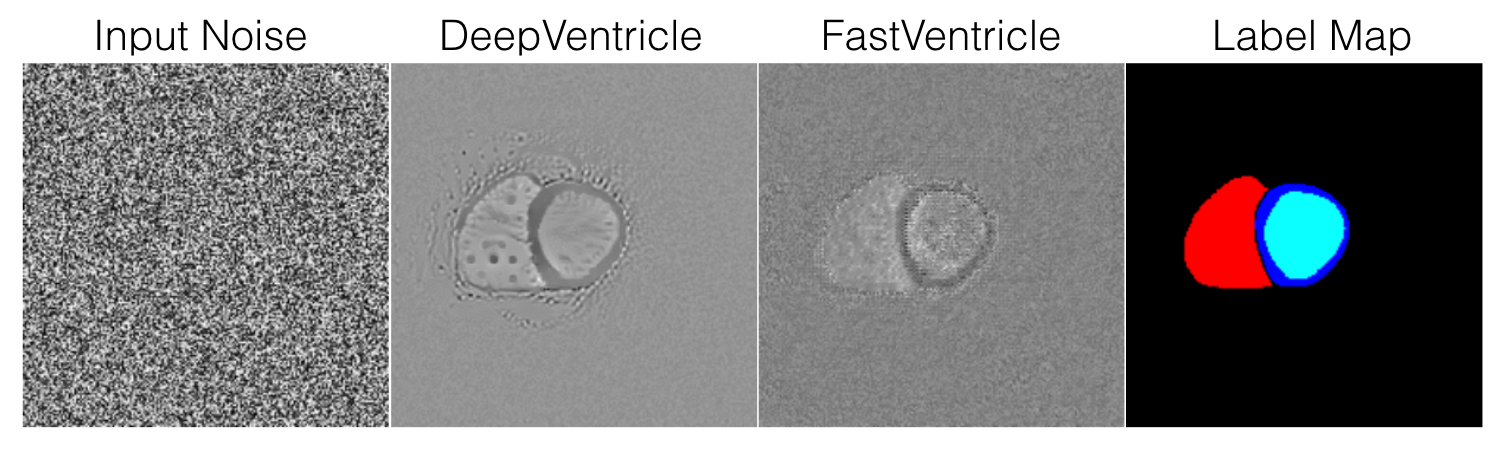}} 
    \\
	\caption{A random input (left) is optimized using gradient descent for DeepVentricle and FastVentricle (middle) to fit the label map (right, RV endo in red, LV endo in cyan, LV epi in blue). The generated image has many qualities that the network is ``looking for" when making its predictions, such as high contrast between endocardium and epicardium and the presence of papillary muscles.}
	\label{fig:deepdream_combined}
\end{figure}

\textbf{Internal Representation.} Neural networks are infamous for being black boxes, i.e., it is very difficult to ``look inside" and understand why a certain prediction is being made. This is especially troublesome in the medical setting, as doctors prefer to use tools that they can understand. We follow the results of \cite{deepdream} to visualize the features that FastVentricle is ``looking for" when performing inference. Beginning with random noise as a model input and a real segmentation mask as the target, we perform backpropagation to update the pixel values in the input image such that the loss is minimized. Figure \ref{fig:deepdream_combined} shows the result of such an optimization for DeepVentricle and FastVentricle. We find that, as a doctor would, the model is confident in its predictions when the endocardium is light and the contrast with the epicardium is high. The model seems to have learned to ignore the anatomy surrounding the heart. We also note that the optimized input for DeepVentricle is less noisy than that for FastVentricle, probably because the former model is larger and utilizes skip connections at the full resolution of the input image. DeepVentricle also seems to ``imagine" structures which look like papillary muscles inside the ventricles.

\section{Discussion}

\textbf{Performance.} Though accuracy is the most important property of a model when making clinical decisions, speed of algorithm execution is also critical for maintaining positive user experience and minimizing infrastructure costs. Within the bounds of our experiments, we find no statistically significant difference between the accuracy of DeepVentricle and that of its $4\times$ faster cousin, FastVentricle. This suggests that FastVentricle can replace DeepVentricle in a clinical setting with no detrimental effects. 

\begin{figure}[tb]
	\centering
    \hspace{\fill} 
    \includegraphics[width = \linewidth]{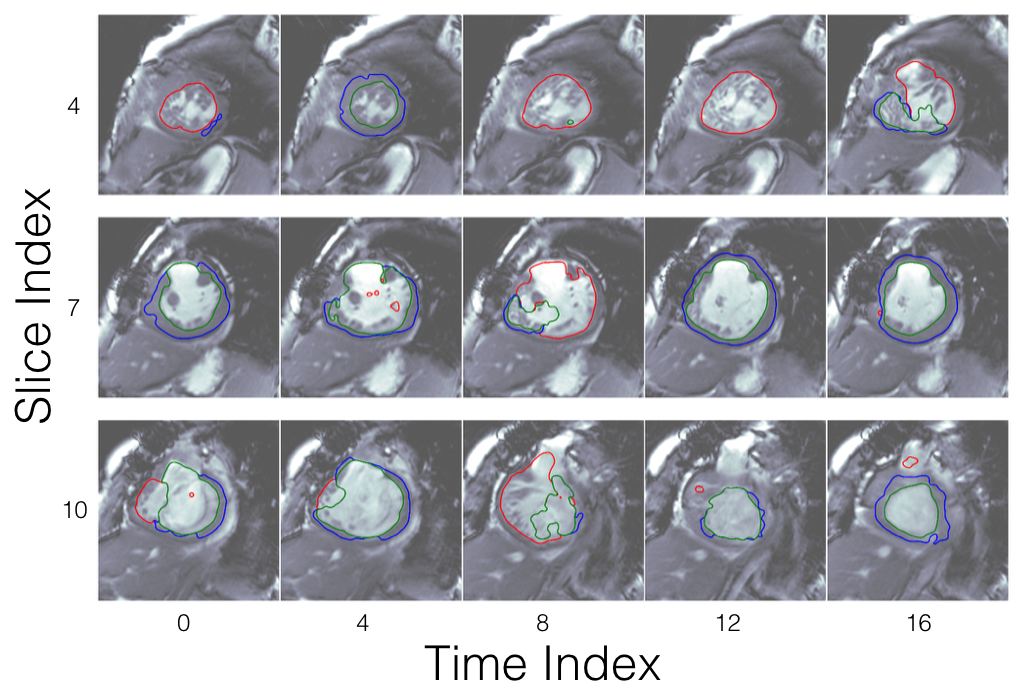}
	\hspace{\fill} 
    \caption{One main failure mode is on patients with congenital defects, for which the anatomy is often ambiguous. In clinical use, a post-processing algorithm that auto-detects poor inference results would report ``cannot find reasonable segmentation" for this study.}
    \label{fig:congenital}
\end{figure}

Both DeepVentricle and FastVentricle have a median RAVE of between $5$ and $7\%$. To put this in context, \cite{suinesiaputra2015quantification} investigated the reproducibility of LV endo volume measurements from CMR over 15 studies with 7 expert readers and found the interrater standard deviation to be, on average, 14mL, i.e. approximately 10$\%$ of the volume being measured. Although there are occasional cases on which our algorithm performs poorly, for example the congenital case displayed in Figure \ref{fig:congenital}, we auto-detect inference results with noisy contours when using DeepVentricle or FastVentricle for clinical use and opt not to display them to the user. Note that this quality detection algorithm has not been run for any plots in this paper; all presented results contain the full hold out set for completeness. For all cases, our production application allows radiologists to view and modify any erroneous contours before completing the report.

\section{Conclusion}
We show that a new ENet-based FCN with skip connections, FastVentricle, can be used for quick and efficient segmentation of cardiac anatomy. Trained on a sparsely annotated database, our algorithm provides LV endo, LV epi, and RV endo contours to clinicians for the purpose of calculating important diagnostic quantities such as ejection fraction and myocardial mass. FastVentricle is $4\times$ faster and runs with $6\times$ less memory than the previous state-of-the-art. 

We experimented with a fully convolutional 3D network to improve the consistency of basal slice segmentations, but found that the memory requirements of this model limited the resolution of the input images such that the results did not outperform 2D segmentation. A more careful consideration of additional spatial information, perhaps with a 2.5D network incorporating long axis views, could potentially improve performance. Finally, as with any deep learning-based model, additional training cases could further improve performance.

\Urlmuskip=0mu plus 1mu\relax
\bibliographystyle{splncs}
\bibliography{biblio}

\begin{thebibliography}{10}

\bibitem{kass1988}
Kass, M., Witkin, A., Terzopoulos, D.:
\newblock Snakes: Active contour models.
\newblock International Journal of Computer Vision (1988)  321--331

\bibitem{choa2006}
Choa, J., Benkeserb, P.J.:
\newblock Cardiac segmentation by a velocity-aided active contour model.
\newblock Computerized Medical Imaging and Graphics \textbf{30} (2006)  31--41

\bibitem{zhu2013geodesic}
Zhu, W.,  et~al.:
\newblock A geodesic-active-contour-based variational model for short-axis
  cardiac {MRI} segmentation.
\newblock Int. Journal of Computer Math. \textbf{90}(1) (2013)

\bibitem{plue2005}
Pluempitiwiriyawej, C.,  et~al.:
\newblock Stacs: New active contour scheme for cardiac {MR} image segmentation.
\newblock In: IEEE Transactions on Medical Imaging. Volume~24. (2005)

\bibitem{schwarz20073d}
Schwarz, T., Heimann, T., Wolf, I., Meinzer, H.:
\newblock 3d heart segmentation and volumetry using deformable shape models.
\newblock In: Computers in Cardiology, 2007, IEEE (2007)  741--744

\bibitem{petitjean2011review}
Petitjean, C., Dacher, J.N.:
\newblock A review of segmentation methods in short axis cardiac mr images.
\newblock Medical image analysis \textbf{15}(2) (2011)  169--184

\bibitem{peng2016review}
Peng, P., Lekadir, K., Gooya, A., Shao, L., Petersen, S.E., Frangi, A.F.:
\newblock A review of heart chamber segmentation for structural and functional
  analysis using cardiac magnetic resonance imaging.
\newblock Magnetic Resonance Materials in Physics, Biology and Medicine
  \textbf{29}(2) (2016)  155--195

\bibitem{long2015fully}
Long, J., Shelhamer, E., Darrell, T.:
\newblock Fully convolutional networks for semantic segmentation.
\newblock In: Proceedings of the IEEE CVPR. (2015)  3431--3440

\bibitem{tran2016fully}
Tran, P.V.:
\newblock A fully convolutional neural network for cardiac segmentation in
  short-axis {MRI}.
\newblock arXiv preprint arXiv:1604.00494 (2016)

\bibitem{noh2015learning}
Noh, H., Hong, S., Han, B.:
\newblock Learning deconvolution network for semantic segmentation.
\newblock In: Proceedings of the IEEE ICCV. (2015)  1520--1528

\bibitem{ronneberger2015u}
Ronneberger, O., Fischer, P., Brox, T.:
\newblock U-net: Convolutional networks for biomedical image segmentation.
\newblock In: MICCAI, Springer (2015)  234--241

\bibitem{lau2016}
Lau, H.K.,  et~al.:
\newblock Deep{Ventricle}: Automated cardiac {MRI} ventricle segmentation using
  deep learning.
\newblock Conference on Machine Intelligence in Medical Imaging (2016)

\bibitem{arterysfda}
Food{\ }and{\ }Drug{\ }Administration:
\newblock Arterys cardio dl.
\newblock \url{http://www.accessdata.fda.gov/cdrh_docs/pdf16/K163253.pdf}
  (January 2017)

\bibitem{paszke2016enet}
Paszke, A., Chaurasia, A.,  et~al.:
\newblock Enet: A deep neural network architecture for real-time semantic
  segmentation.
\newblock arXiv preprint arXiv:1606.02147 (2016)

\bibitem{he2016deep}
He, K., Zhang, X., Ren, S., Sun, J.:
\newblock Deep residual learning for image recognition.
\newblock In: Proceedings of the IEEE CVPR. (2016)  770--778

\bibitem{yu2015multi}
Yu, F., Koltun, V.:
\newblock Multi-scale context aggregation by dilated convolutions.
\newblock arXiv preprint arXiv:1511.07122 (2015)

\bibitem{chollet2015keras}
Chollet, F.:
\newblock Keras.
\newblock \url{https://github.com/fchollet/keras} (2015)

\bibitem{abadi2016tensorflow}
Abadi, M.,  et~al.:
\newblock Tensorflow: Large-scale machine learning on heterogeneous distributed
  systems.
\newblock arXiv preprint arXiv:1603.04467 (2016)

\bibitem{kingma2014adam}
Kingma, D., Ba, J.:
\newblock Adam: A method for stochastic optimization.
\newblock arXiv preprint arXiv:1412.6980 (2014)

\bibitem{bergstra2012random}
Bergstra, J., Bengio, Y.:
\newblock Random search for hyper-parameter optimization.
\newblock Journal of Machine Learning Research \textbf{13}(Feb) (2012)
  281--305

\bibitem{bland1986statistical}
Bland, J.M., Altman, D.:
\newblock Statistical methods for assessing agreement between two methods of
  clinical measurement.
\newblock The lancet \textbf{327}(8476) (1986)  307--310

\bibitem{suinesiaputra2015quantification}
Suinesiaputra, A.,  et~al.:
\newblock Quantification of lv function and mass by cardiovascular magnetic
  resonance: multi-center variability and consensus contours.
\newblock Journal of Cardiovascular Magnetic Resonance \textbf{17}(1) (2015)
  ~63

\bibitem{deepdream}
Mordvintsev, A.,  et~al.:
\newblock Deep {D}ream.
\newblock
  \url{https://research.googleblog.com/2015/06/inceptionism-going-deeper-into-neural.html}
  (2015) Accessed: 2017-01-17.

\end{thebibliography}

\end{document}